\documentclass{article}
\usepackage{spconf,amsmath,graphicx,hyperref}
\usepackage{booktabs}
\title{COLLAPSE, NOT COMPLEXITY: FAILURE-CONDITIONED DECOMPOSITION REPAIR \\ FOR END-TO-END DOCUMENT PARSING}

\name{Xingyu Lin, Dehui Du}
\address{Software Engineering Institute, East China Normal University \\ Shanghai, China \\ xingyulin@stu.ecnu.edu.cn, dhdu@sei.ecnu.edu.cn}

\begin{document}
\frenchspacing
\ninept
\maketitle

\begin{abstract}
End-to-end document parsers increasingly offer an optional reasoning mode for complex
pages. On a 180-page entropy-stratified discovery sample with one frozen 4B checkpoint,
complexity is the wrong decision variable. Reasoning lowers mean quality by 2.21 Overall at
$1.54\times$ tokens; a preregistered input-only model cannot predict its signed benefit
(held-out AUROC $0.47$, indistinguishable from chance). The benefit concentrates on pages
whose ordinary pass has already collapsed, and they do not look complex: shared
collapses have \emph{lower} layout entropy than healthy ones yet consume $19\times$ the
tokens as degenerate repetition that doubling the budget does not cure. Switching modes
rarely repairs them: $83\%$ recur under reasoning. We instead detect collapse from the
ordinary-pass trace, decompose the page by projection, and re-parse each region. Repair
gains $1.40$ Overall (95\% CI $[0.68, 2.16]$) at $1.13\times$ tokens, replicates across
three checkpoints, and, with all parameters frozen, gains $2.41$ (CI $[1.64,
3.46]$) on the remaining $1{,}175$ benchmark pages.
\end{abstract}


\section{Introduction}
\label{sec:intro}

End-to-end document vision-language models (VLMs) map a page image directly to Markdown,
replacing cascades of layout detection, text recognition and structure
reconstruction~\cite{qianfanocr,paddleocrvl,mineru25,olmocr}. Because a single
autoregressive pass must emit the whole page, these models inherit a failure mode that
conventional cascades largely avoid: the decoder can exhaust its token budget and return
nothing usable. Several such parsers now ship an \emph{optional reasoning mode} (an
intermediate trace emitted before the final Markdown), and vendor guidance recommends
it for structurally complex pages~\cite{qianfanocr}, echoing input-dependent reasoning
or budget selection~\cite{adaptthink,zerostep,thinkornot,movt,gpro,certaintyrouting}.
The switch lengthens every output it touches, so a wrong decision costs
quality and compute at once.

We therefore ask: \emph{what actually decides whether the reasoning mode helps a
document page?} The prevailing framing, judging how complex the page looks before running
anything, picks the wrong variable. Conditioning on observed failure instead turns an
unpromising mode-selection problem into a tractable repair problem.

\noindent Our contributions are:
\begin{enumerate}
\itemsep0.10em
\item A paired diagnosis: reasoning is harmful on average on the stratified sample;
pre-inference features cannot predict its signed benefit, which concentrates on
ordinary-pass collapse (Sec.~\ref{sec:diagnosis}).
\item A design consequence: runaway generation dominates cost, while $83\%$ of ordinary
collapses persist under reasoning.
\item A reference-free trigger followed by decomposition and re-parsing: $+1.40$ Overall
at $1.13\times$ tokens on the discovery set, within $0.36$ points of the gold-label oracle;
it replicates across three backbones spanning 3B--8B and, with all parameters frozen,
gains $+2.41$ (CI $[1.64, 3.46]$) on $1{,}175$ held-out benchmark pages, with explicit
indivisible-block and figure-dominated failure cases (Sec.~\ref{sec:exp}).
\end{enumerate}

\begin{figure}[t]
\centering
\includegraphics[width=\columnwidth]{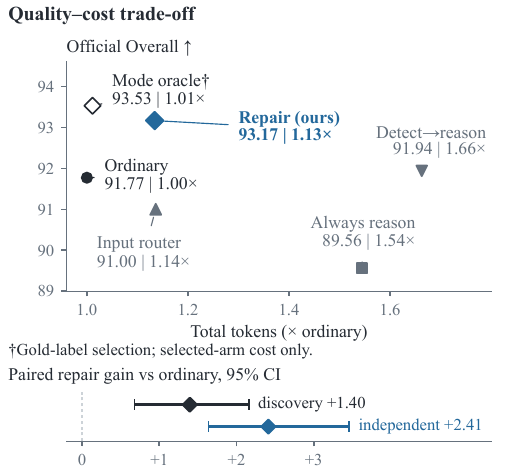}
\caption{Quality--cost comparison on 180 OmniDocBench v1.5 pages, one frozen 4B checkpoint.
Labels give Overall and relative token cost. $^\dagger$The non-deployable oracle assumes
free gold-label selection and charges only the chosen arm. Detect$\rightarrow$reason re-runs flagged
pages with reasoning, charging both calls. Repair pays for the ordinary pass and all
region calls, including failures. The lower panel reports paired repair gains and their
95\% intervals on the 180-page discovery set and the 1{,}175-page independent set, not
absolute-score uncertainty.}
\label{fig:framework}
\end{figure}

\section{Relation to prior work}
\label{sec:prior}

\noindent\textbf{Reasoning-mode selection.} AdaptThink~\cite{adaptthink}, Zero-Step
Thinking~\cite{zerostep}, selective reasoning for VLMs~\cite{thinkornot},
Mixture-of-Visual-Thoughts~\cite{movt}, gated perception--reasoning
optimisation~\cite{gpro} and certainty-based adaptive routing~\cite{certaintyrouting} all
choose a reasoning mode, budget or strategy, typically by training the policy or reading
internal states, and are evaluated on general reasoning or VQA. We leave the
checkpoint frozen and instead target failures that persist across modes. When reasoning
hurts, the effect is established for tasks where deliberation harms
humans~\cite{mindyourstep} and for perception-heavy multimodal
tasks~\cite{looklight}; format-decoupled training likewise separates reading from
reasoning in document OCR~\cite{fdrl}. We localise the effect in full-page parsing
under official structural metrics, explain it as a repetition pathology, and supply a
repair rather than a diagnosis alone.

\noindent\textbf{Routing for document models.} Predicting OCR accuracy from simple image
features dates to ICDAR 1995~\cite{ocracc1995}; recent document routing runs OCR first and
then decides~\cite{preinferrouting}, or routes by visual style to domain
experts~\cite{manchu}. We quantify the gap directly: for the failure target, pre-inference
page features are substantially weaker than one post-hoc output statistic
(Sec.~\ref{ssec:detector}). Reference-free quality estimation is related but heavier:
Consensus Entropy~\cite{consensusentropy} needs agreement among several VLMs and
DOCR-Inspector~\cite{docrinspector} trains a separate evaluator, whereas our trigger reads
only the already-paid decoding trace. LayoutLite~\cite{layoutlite}, PaddleOCR-VL~\cite{paddleocrvl}
and MinerU2.5~\cite{mineru25} prune visual tokens or adopt coarse-to-fine architectures;
these redesign the model, whereas our intervention follows observed failure without any
retraining. PureDocBench~\cite{puredocbench} audits OmniDocBench~\cite{omnidocbench}
annotation quality; our analysis is orthogonal, targeting cost--quality and reasoning-mode
effects on the same benchmark. Projection /
X-Y cut~\cite{xycut} and Voronoi~\cite{voronoi} page segmentation
are classical; we claim no novelty in segmentation itself, only in triggering it
conditionally on observed failure.

\begin{figure*}[t]
\centering
\includegraphics[width=\textwidth]{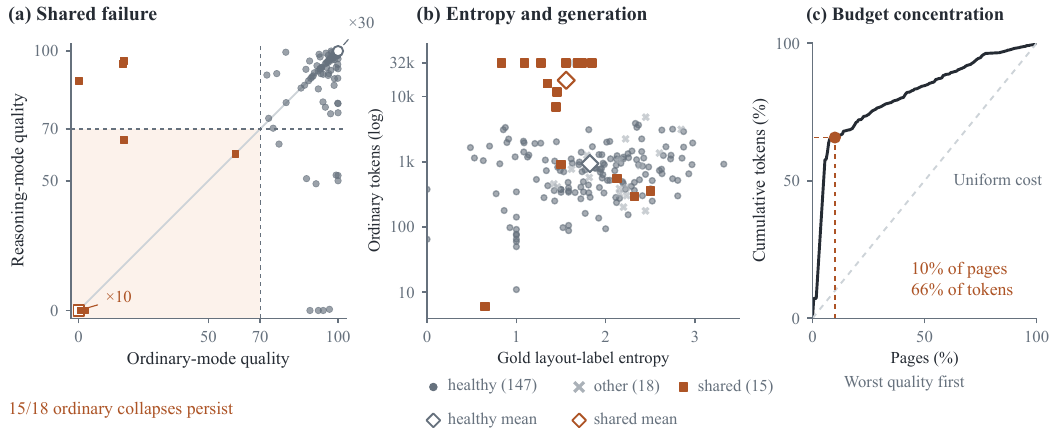}
\caption{Why repair targets observed failure. (a) Paired ordinary/reasoning page quality;
orange squares have ordinary quality $<70$; dashed lines mark $70$; the diagonal denotes
equal scores; exact overlaps are counted without jitter. The shaded region contains 15 of
the 18 ordinary collapses. (b) All 180 pages in entropy--token space (log token axis):
healthy means ordinary quality $\geq90$; shared means both modes $<70$; group diamonds
show descriptive means. Entropy is a complexity proxy. (c) Ordinary-mode token
concentration, ranked by increasing ordinary quality.}
\label{fig:evidence}
\end{figure*}

\section{Layout complexity is the wrong variable}
\label{sec:diagnosis}

\subsection{Protocol}
\label{ssec:protocol}
We sample 180 pages from OmniDocBench v1.5~\cite{omnidocbench} (exactly 20 from each
of its 9 document genres, stratified into thirds by gold layout-label entropy before any
inference) and run one frozen 4B document VLM~\cite{qianfanocr} twice per page. The
\emph{only} manipulated variable is the reasoning-mode switch; prompt, image, random seed
and greedy decoding are identical across arms. Quality uses the official component
metrics (text edit distance, table
TEDS~\cite{teds} and formula CDM) and the official subset Overall, their mean.
Generation failures score zero quality at full token cost and are never dropped, because
dropping them would hide the very phenomenon we study. Throughout, we call a page
\emph{collapsed} when its ordinary pass scores below $70$ and \emph{healthy} when it
scores at least $90$. Uncertainty is a document-group cluster bootstrap with $10^4$
resamples that recomputes the official aggregate on every resample. The protocol,
including the detector rule and thresholds of
Sec.~\ref{sec:method}, was frozen and hashed before the repair experiments ran.

\subsection{Reasoning is harmful on average}
\label{ssec:harm}
Overall falls from $91.77$ to $89.56$ ($-2.21$) while generated tokens rise to
$1.54\times$ (Fig.~\ref{fig:framework}). Behind this aggregate sits a strong asymmetry:
of the $178$ pages with a supported paired score, $21.4\%$ lose at least one Overall point
when reasoning is enabled, whereas only $9.6\%$ gain that much. The harm concentrates by
genre on dense multi-column newspaper pages, precisely where the ordinary pass
collapses (Sec.~\ref{ssec:volume}). This average harm does not hold out of sample: on a
preregistered held-out subsample it disappears (Sec.~\ref{ssec:independent}).

\subsection{The benefit is not predictable before inference}
A preregistered leave-one-genre-out estimator over raw-pixel projection and whitespace
features predicts the \emph{signed} per-page benefit at AUROC $0.472$ (95\% CI $[0.33,
0.65]$), statistically indistinguishable from chance. We find that these
pre-inference appearance features cannot route reasoning by benefit. This result does not
imply that appearance is uninformative about the distinct target of collapse, which we
evaluate in Sec.~\ref{ssec:detector}.

\subsection{The benefit is conditioned on collapse}
Fig.~\ref{fig:evidence}(a) shows the paired outcomes. Reasoning gains
$+16.1$ Overall points on average on collapsed pages, while on healthy pages it
\emph{loses} $4.5$, a reversal that single-number reporting completely hides.
Collapsed pages are only $10\%$ of the sample, yet they carry $75.5\%$ of all positive
benefit mass; under an independent reference-free split that never reads quality, the
flagged stratum still shows a $+8.7$ mean benefit versus $-3.1$ for the unflagged stratum,
and the observed collapsed-versus-healthy difference of $+20.6$ exceeds the $[5.4, 17.7]$
no-effect null band, so we treat the concentration as a descriptive anchor rather than a
precise causal share.

\subsection{Collapse is volume, not complexity}
\label{ssec:volume}
Pages that collapse in \emph{both} modes have \emph{lower} gold layout entropy than
healthy pages ($1.56$ vs.\ $1.82$) yet consume about $19\times$ as many generated tokens
($17.7$k vs.\ $0.9$k; terminal failures are charged over two capped attempts at the
$16{,}384$-token cap, so counts reach $32{,}768$; Fig.~\ref{fig:evidence}(b)). In shared
collapses the decoder either runs away to its cap or stops after almost no usable text. Thirteen pages fail terminally in
at least one arm, twelve of them newspaper. Doubling
the cap to $32{,}768$ does not repair them: six of ten re-run pages still exhaust the
doubled budget, and all ten outputs are degenerate repetition (median gzip compression
ratio $70$ versus ${\sim}2$ healthy; distinct 10-gram rate $0.019$ versus ${\sim}0.90$).
Collapse is a repetition pathology, not a difficulty of reading a visually intricate
layout.

The worst $10\%$ of pages absorb $66\%$ of the entire
token budget (Fig.~\ref{fig:evidence}(c)).

\subsection{Why mode selection cannot fix it}
Executed as a deployable cascade (both calls charged),
this detect-then-reason baseline gains only $+0.17$ Overall ($91.94$) at $1.66\times$
tokens, the highest cost of any policy in Fig.~\ref{fig:framework}; reasoning improves
only one of the ten flagged pages by more than a point. The deeper reason is that $15$ of
the $18$ collapsed pages ($83\%$) collapse under reasoning as well. Consistently, a
hindsight oracle that always picks the better mode using gold labels reaches only $93.53$
Overall (its $1.01\times$ cost counts only the chosen arm, assuming free perfect
selection). Because most collapses persist across modes, the mode switch is the wrong
control variable, motivating a different intervention.

\section{Failure-conditioned decomposition repair}
\label{sec:method}

\subsection{Detecting collapse from the output side}
\label{ssec:detector}
Our detector reads only the decoding trace of the pass we have already paid for. Let
$n(x)$ be the number of tokens that pass emitted on page $x$, and let $s(x)$ be its
termination status, which is either a normal stop or a run to the token cap $T$. We flag
the page when
\begin{equation}
\label{eq:detector}
\mathrm{flag}(x) \;=\; \mathbf{1}\!\left[\, s(x) \neq \textit{stop} \;\;\vee\;\; n(x) \geq \tau \,\right].
\end{equation}
The frozen operating point is $\tau = 8000$, about half the $T = 16{,}384$-token cap.
At this operating point the trigger recalls $10$ of the $18$ discovery-set collapses
($55.6\%$) while flagging no healthy page; on the independent test set of
Sec.~\ref{ssec:independent} it recalls $70.6\%$ at a $0.11\%$ healthy false-positive
rate. Table~\ref{tab:detector} ablates $\tau$.

Page appearance still carries some failure signal for the distinct target of
ordinary-pass collapse: pre-inference features reach AUROC $0.679$, the training-free
statistic \emph{characters per generated token} $0.879$. The frozen rule is not dominated:
on the independent set a chars-per-token trigger costs zero healthy false positives at
$64\%$ recall, but $10\%$ at the frozen rule's $70.6\%$ recall.

\begin{table}[t]
\centering
\caption{Ablation of the detector threshold $\tau$ of Eq.~\eqref{eq:detector} on the
180-page sample. ``True fail'' counts flagged pages that collapsed ($<70$), ``False
pos.'' flagged healthy pages ($\geq 90$), ``Budget'' the flagged share of all generated
tokens. No setting in the preregistered envelope flags a healthy page; the
$\geq\!8000$ row is the frozen operating point.}
\label{tab:detector}
\small
\begin{tabular}{@{}lrrrr@{}}
\toprule
Trigger & Flagged & True fail & False pos.\ & Budget \\
\midrule
no stop token only      & 8  & 8  & 0 & 57.5\% \\
$\geq\!16000$ tokens    & 8  & 8  & 0 & 57.5\% \\
$\geq\!8000$ tokens     & 10 & 10 & 0 & 63.6\% \\
$\geq\!4000$ tokens     & 12 & 11 & 0 & 66.2\% \\
\bottomrule
\end{tabular}
\end{table}

\subsection{Repair by decomposition}
A flagged page is decomposed by projection segmentation with fixed gap parameters into
regions ordered top-to-bottom then left-to-right; each region crop is parsed by the
\emph{same} frozen checkpoint in the \emph{same} ordinary mode, and the region outputs
are concatenated in that reading order. The only variable changing is input granularity,
isolating decomposition from reasoning-mode or prompt confounds. Resampling is not an
alternative: decoding is greedy with a fixed seed, so
re-running a collapsed page reproduces it byte for byte. An unflagged page retains its
original output with no extra model call, and a flagged page yielding fewer than two
regions is left unchanged, so non-repairable failures cannot quietly leave the
denominator.

The trigger is load-bearing, not decorative. Decomposing \emph{every} page with the same
segmentation collapses Overall to $76.36$ ($-15.41$): it destroys
table structure (TEDS $93.4\rightarrow 53.9$) and makes $35$ healthy pages
worse while helping only $5$, a $16.8$-point swing attributable to the
trigger alone.

\begin{figure}[t]
\centering
\includegraphics[width=\columnwidth]{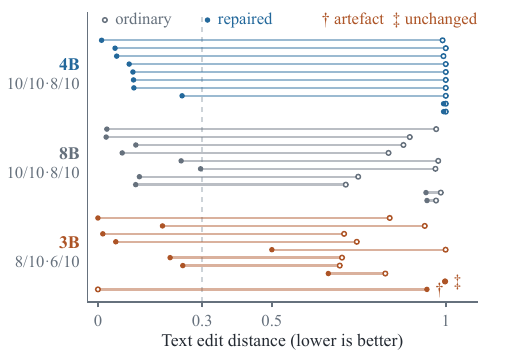}
\caption{Grouped paired dot plot of per-page repair outcomes (lower is better). Each row
is one frozen flagged page: the open dot is its ordinary-pass text edit distance, the
filled dot the repaired distance, so a leftward span is an improvement. Colour identifies
the checkpoint (4B/8B repair the 4B flagged set; 3B its own). Group labels give pages
improved $\cdot$ pages reaching $\leq\!0.3$; the dashed line marks that tally threshold,
not a significance level. All 30 pages appear, including 4B residual failures, 8B
high-error small improvements, the unchanged page ($\ddagger$) and the 3B metric artefact
($\dagger$) --- a measurement degeneration, not a real regression. Table~\ref{tab:cross}
gives the official Overall gains under each checkpoint's own trigger.}
\label{fig:repair}
\end{figure}

\section{Experiments}
\label{sec:exp}
We reuse the frozen protocol of Sec.~\ref{ssec:protocol} unchanged (the same 180
OmniDocBench v1.5 pages, the same official metrics and the same document-group cluster
bootstrap), so any quality difference is attributable to the policy rather than to
sampling variation.

\subsection{Main result}
Decomposition repair raises Overall from $91.77$
to $93.17$ ($+1.396$; 95\% CI $[+0.68, +2.16]$) at $1.13\times$ tokens, the highest
Overall among the evaluated deployable policies, $0.36$ behind the hindsight oracle and
$2.17$ above the frozen input router at essentially the same cost.

On the flagged stratum, mean text edit distance falls by $0.725$; eight of ten pages drop
to $0.01$--$0.24$ (Fig.~\ref{fig:repair}). The repair spent 55 region calls, two of which
hit the cap themselves and are charged in full.

Tokens are not the only cost: wall-clock rises to $1.31\times$ here and
$1.25\times$ on the independent set, above the token multiples (each crop is
re-encoded) yet below always-reason's $1.58\times$; the vision side adds only
$1.12\times$ pixels because regions \emph{tile} the flagged page rather than duplicate it;
wall-clock is deployed shared-server cost, not a controlled benchmark.

\subsection{Cross-checkpoint replication}
We repeat the pipeline on two further open-weight parsers
(Table~\ref{tab:cross}; Fig.~\ref{fig:repair}), each using \emph{its own} trigger on
\emph{its own} failures, with the same official evaluator. All three
checkpoints collapse on dense pages and benefit from repair; the failure \emph{form}
differs while the fix does not. The trigger fires at similar rates (4B and 3B:
$5.6\%$, 8B: $6.7\%$). The 8B also reproduces the diagnosis independently: its
reasoning mode triples the runaway rate ($6.7\%\rightarrow22.8\%$).

\begin{table}[t]
\centering
\caption{Self-detect--self-repair with each checkpoint's own trigger, official evaluation
(with CDM) on the full 180 pages. All rows are all-inclusive; flagged-but-indivisible
pages keep their ordinary output and no page is dropped.}
\label{tab:cross}
\small
\begin{tabular}{@{}lrrrr@{}}
\toprule
Checkpoint & Flagged & Baseline & Self-repair & $\Delta$ \\
\midrule
4B~\cite{qianfanocr}  & 10/180 & 91.77 & 93.17 & $+1.40$ \\
8B~\cite{minicpmv45}  & 12/180 & 83.35 & 84.49 & $+1.14$ \\
3B~\cite{nanonetsocr2}& 10/180 & 81.18 & 81.88 & $+0.70$ \\
\bottomrule
\end{tabular}
\end{table}

\subsection{Independent test set}
\label{ssec:independent}
The discovery sample supported analysis, design and evaluation simultaneously,
so we froze every parameter and re-ran the pipeline on the remaining $1{,}175$
OmniDocBench v1.5 pages, the exhaustive page-disjoint complement with nothing tuned on it
(60 of the 163 discovery document groups reappear; a post-hoc restriction to the 611 pages
whose documents never appear in discovery still gains $+3.35$, CI $[+2.13, +4.89]$). The natural distribution is harder (baseline
Overall $88.05$) and collapses more often ($10.8\%$), so the trigger fires on $7.8\%$ of
pages, recalling $89$ of $126$ collapses ($70.6\%$) with one healthy false positive in
$946$. Repair raises Overall to $90.46$, a paired gain of $+2.41$ (95\%
CI $[+1.64, +3.46]$) at $1.18\times$ tokens, \emph{larger} than on the discovery set:
the effect is not an artefact of the stratified sample. On the flagged stratum, $69$ of
the $92$ pages improve by more than a point, $21$ are unchanged and $2$ degrade; $57$
gain more than $50$ points. Dropping all $131$ newspaper pages still leaves $+1.02$
(CI $[+0.31, +1.87]$); $7$ of $9$ genres improve and none is harmed. The single healthy
false positive (an exam page completing at $8{,}810$ tokens, just above $\tau$) is
repaired from $95.7$ to $84.5$, the boundary cost of the operating point.

The reasoning-mode diagnosis also holds out of sample: on a preregistered $200$-page
subsample reasoning does \emph{not} lower Overall ($\Delta{+}2.25$, CI $[-0.39,+5.90]$),
so the average harm of Sec.~\ref{ssec:harm} is a property of the stratified discovery
sample, yet its mechanism replicates ($+14.1$ on collapsed vs.\ $-3.5$ on healthy
pages; $74\%$ of collapses recur), and detect-then-reason again trails repair ($+0.66$ at
$1.69\times$ tokens).

\subsection{Limitations: where the repair does not work}
\label{ssec:boundaries}

\noindent\textbf{Decomposition helps only when the failure decomposes.} The two unrepaired
4B pages each localise their runaway to one indivisible dense block. Segmentation returns a
single region, so the repair correctly declines to act. On the independent set,
9 of the 92 flagged pages fall in this category and are left unchanged. Such pages
require a hard budget or a segmentation prior not reliant on whitespace.

\noindent\textbf{``Used most of the budget'' is not synonymous with failure.} On one
figure-dominated page with 19 characters of gold text, both outputs hallucinate heavily;
the baseline happens to begin with the exact gold and scores perfectly, making repair
appear harmful. The independent-set exam page of Sec.~\ref{ssec:independent} is a second,
milder instance. The Table~\ref{tab:cross} deltas keep both pages; excluding the 3B
artefact moves its stratum mean from $+0.36$ to $+0.51$ on the old edit-distance measure
--- a sensitivity, not a correction.

\noindent\textbf{The method presumes an end-to-end parser.} A region-level
recogniser designed to follow layout analysis~\cite{glmocr} is out of scope.

\section{Conclusion}
For end-to-end document parsing, whether an optional reasoning mode helps a page is decided
not by how complex the page looks but by whether the ordinary pass has already collapsed
--- and collapse is a degenerate-repetition pathology that doubling the budget does not
cure and the reasoning mode largely shares. Conditioning on observed failure rather than
predicted difficulty turns ineffective mode selection into repair: $+1.40$ Overall at
$1.13\times$ cost, replicated across three checkpoints, and $+2.41$ (CI $[1.64, 3.46]$) on
$1{,}175$ held-out pages with every parameter frozen.

\section{Compliance with ethical standards}
Only the public OmniDocBench v1.5 benchmark~\cite{omnidocbench} and public model
checkpoints are used; no human subjects, personal data or private corpora are involved.

\section{Acknowledgements}
No specific funding was received. Code, per-page outputs and hashed protocols will be
released upon acceptance. Per the ICASSP 2027 LLM policy: an AI assistant supported
language editing, analysis/plotting code and literature search; all experiments, numbers,
code and claims were verified by the author; no section was produced wholesale by an LLM.

\frenchspacing
\let\spconfthebibliography\thebibliography
\def\thebibliography#1{\spconfthebibliography{#1}\def\newblock{\hskip .11em\relax}}
\bibliographystyle{IEEEbib}
\bibliography{refs}

\begin{thebibliography}{10}

\bibitem{qianfanocr}
{Baidu Qianfan Team},
\newblock ``{Qianfan-OCR}: A unified end-to-end model for document
  intelligence,''
\newblock Tech. {R}ep., Baidu, 2026,
\newblock arXiv:2603.13398.

\bibitem{paddleocrvl}
Cheng Cui, Ting Sun, Suyin Liang, Tingquan Gao, Zelun Zhang, et~al.,
\newblock ``{PaddleOCR-VL}: Boosting multilingual document parsing via a 0.9{B}
  ultra-compact vision-language model,''
\newblock {\em arXiv:2510.14528}, 2025.

\bibitem{mineru25}
Junbo Niu, Zheng Liu, Zhuangcheng Gu, Bin Wang, Linke Ouyang, et~al.,
\newblock ``{MinerU2.5}: A decoupled vision-language model for efficient
  high-resolution document parsing,''
\newblock in {\em Proc. Annual Meeting of the Association for Computational
  Linguistics (ACL), Industry Track}, 2026.

\bibitem{olmocr}
Jake Poznanski et~al.,
\newblock ``The {olmOCR} project: Building fully open {OCR} using {VLMs},''
\newblock in {\em Proc. Annual Meeting of the Association for Computational
  Linguistics (ACL), System Demonstrations}, 2026.

\bibitem{adaptthink}
Jiajie Zhang, Nianyi Lin, Lei Hou, Ling Feng, and Juanzi Li,
\newblock ``{AdaptThink}: Reasoning models can learn when to think,''
\newblock in {\em Proc. EMNLP}, 2025.

\bibitem{zerostep}
Yuqiao Tan, Shizhu He, Kang Liu, and Jun Zhao,
\newblock ``The zero-step thinking: An empirical study of mode selection as
  harder early exit in reasoning models,''
\newblock in {\em NeurIPS Workshop on Efficient Reasoning}, 2025.

\bibitem{thinkornot}
Jiaqi Wang, Kevin~Qinghong Lin, James Cheng, and Mike~Zheng Shou,
\newblock ``Think or not? selective reasoning via reinforcement learning for
  vision-language models,''
\newblock in {\em Proc. Advances in Neural Information Processing Systems
  (NeurIPS)}, 2025.

\bibitem{movt}
Zejun Li, Yingxiu Zhao, Jiwen Zhang, Siyuan Wang, Yang Yao, et~al.,
\newblock ``Mixture-of-visual-thoughts: Exploring context-adaptive reasoning
  mode selection for general visual reasoning,''
\newblock in {\em Proc. International Conference on Learning Representations
  (ICLR)}, 2026.

\bibitem{gpro}
Xingjian Diao, Zheyuan Liu, Chunhui Zhang, Weiyi Wu, Keyi Kong, et~al.,
\newblock ``Addressing overthinking in large vision-language models via gated
  perception-reasoning optimization,''
\newblock in {\em Findings of the Association for Computational Linguistics
  (ACL)}, 2026.

\bibitem{certaintyrouting}
Jinghui Lu, Haiyang Yu, Siliang Xu, Shiwei Ran, GuoZhi Tang, et~al.,
\newblock ``Prolonged reasoning is not all you need: Certainty-based adaptive
  routing for efficient {LLM/MLLM} reasoning,''
\newblock {\em arXiv:2505.15154}, 2025.

\bibitem{mindyourstep}
Ryan Liu, Jiayi Geng, Addison~J. Wu, Ilia Sucholutsky, Tania Lombrozo, and
  Thomas~L. Griffiths,
\newblock ``Mind your step (by step): Chain-of-thought can reduce performance
  on tasks where thinking makes humans worse,''
\newblock in {\em Proc. ICML}, 2025.

\bibitem{looklight}
Zhuoran Jin, Kejian Zhu, Hongbang Yuan, Yupu Hao, Pengfei Cao, Yubo Chen, Kang
  Liu, and Jun Zhao,
\newblock ``Look light, think heavy: What multimodal chain-of-thought reasoning
  can and cannot do,''
\newblock in {\em Proc. Annual Meeting of the Association for Computational
  Linguistics (ACL)}, 2026.

\bibitem{fdrl}
Yufeng Zhong, Lei Chen, Zhixiong Zeng, Xuanle Zhao, Deyang Jiang, et~al.,
\newblock ``Reading or reasoning? format decoupled reinforcement learning for
  document {OCR},''
\newblock {\em arXiv:2601.08834}, 2025.

\bibitem{ocracc1995}
Luis~R. Blando, Junichi Kanai, and Thomas~A. Nartker,
\newblock ``Prediction of {OCR} accuracy using simple image features,''
\newblock in {\em Proc. International Conference on Document Analysis and
  Recognition (ICDAR)}, 1995, pp. 319--322.

\bibitem{preinferrouting}
Sreerekha Rajendran,
\newblock ``Pre-inference routing for cost-efficient document field
  extraction,''
\newblock {\em arXiv:2608.06607}, 2026.

\bibitem{manchu}
Zhan Chen, Jiqiao Ma, and Chih-wen Kuo,
\newblock ``Multi-expert routing for multi-domain low-resource {OCR}: A
  {Manchu} case study,''
\newblock {\em arXiv:2607.14041}, 2026.

\bibitem{consensusentropy}
Yulong Zhang, Tianyi Liang, Erfei Cui, Guoqing Wang, et~al.,
\newblock ``Consensus entropy: Harnessing multi-{VLM} agreement for
  self-verifying and self-improving {OCR},''
\newblock in {\em Proc. IEEE/CVF Conf. Computer Vision and Pattern Recognition
  (CVPR)}, 2026.

\bibitem{docrinspector}
Qintong Zhang, Junyuan Zhang, Zhifei Ren, Linke Ouyang, Zichen Wen, et~al.,
\newblock ``{DOCR-Inspector}: Fine-grained and automated evaluation of document
  parsing with {VLM},''
\newblock {\em arXiv:2512.10619}, 2025.

\bibitem{layoutlite}
Xudong Liu, Bicheng Wan, and Yulin Jin,
\newblock ``{LayoutLite}: Token-level implicit layout analysis for efficient
  document {OCR},''
\newblock {\em arXiv:2607.22200}, 2026.

\bibitem{puredocbench}
Zhiheng Li, Zongyang Ma, Jiaxian Chen, Jianing Zhang, Zhaolong Su, et~al.,
\newblock ``How far is document parsing from solved? {PureDocBench}: A
  source-traceable benchmark across clean, degraded, and real-world settings,''
\newblock {\em arXiv:2605.07492}, 2026.

\bibitem{omnidocbench}
Linke Ouyang, Yuan Qu, Hongbin Zhou, Jiawei Zhu, Rui Zhang, et~al.,
\newblock ``{OmniDocBench}: Benchmarking diverse {PDF} document parsing with
  comprehensive annotations,''
\newblock in {\em Proc. IEEE/CVF Conf. Computer Vision and Pattern Recognition
  (CVPR)}, 2025.

\bibitem{xycut}
George Nagy and Sharad Seth,
\newblock ``Hierarchical representation of optically scanned documents,''
\newblock in {\em Proc. International Conference on Pattern Recognition
  (ICPR)}, 1984, pp. 347--349.

\bibitem{voronoi}
Koichi Kise, Akinori Sato, and Motoi Iwata,
\newblock ``Segmentation of page images using the area {V}oronoi diagram,''
\newblock {\em Computer Vision and Image Understanding}, vol. 70, no. 3, pp.
  370--382, 1998.

\bibitem{teds}
Xu~Zhong, Elaheh ShafieiBavani, and Antonio Jimeno~Yepes,
\newblock ``Image-based table recognition: Data, model, and evaluation,''
\newblock in {\em Proc. European Conference on Computer Vision (ECCV)}, 2020.

\bibitem{minicpmv45}
{OpenBMB},
\newblock ``{MiniCPM-V} 4.5: Cooking efficient {MLLMs} via architecture, data,
  and training recipes,''
\newblock Tech. {R}ep., OpenBMB, 2025,
\newblock arXiv:2509.18154.

\bibitem{nanonetsocr2}
{Nanonets},
\newblock ``{Nanonets-OCR2}: Transforming documents into structured markdown,''
  Model card, nanonets/Nanonets-OCR2-3B, 2025.

\bibitem{glmocr}
Shuaiqi Duan, Yadong Xue, Weihan Wang, Zhe Su, Huan Liu, et~al.,
\newblock ``{GLM-OCR} technical report,''
\newblock Tech. {R}ep., Zhipu AI, 2026,
\newblock arXiv:2603.10910.

\end{thebibliography}

\end{document}